\documentclass[letterpaper]{article}
\usepackage{authblk}
\usepackage[utf8]{inputenc}
\usepackage[numbers]{natbib}
\usepackage{graphicx}
\usepackage{longtable}
\usepackage{amsfonts, amsthm, amsmath, amssymb}
\usepackage{mathrsfs}
\usepackage{graphicx}
\usepackage{color}
\usepackage{tabularx}
\usepackage{enumerate}
\usepackage{algorithm, algorithmic}
\usepackage{xspace}
\usepackage{algorithm, algorithmic}
\usepackage{tikz}
\usepackage{multirow}
\usepackage{makecell}
\usepackage{booktabs}
\usepackage{pdflscape}
\usepackage{geometry}
\usepackage{setspace}
\usepackage{multicol}
\usepackage{supertabular}
\usepackage{hyperref}
\usepackage{pdflscape}

\usepackage{float}
\usepackage{authblk}
\usepackage{siunitx}

\usepackage{amssymb}
\usepackage{amsmath}
\usepackage{tikz}
\usepackage{helvet} 
\usepackage{courier} 
\title{A novel hallucination classification framework}
\author{Maksym Zavhorodnii \and Dmytro Dehtiarov\\ \textit{ Instituto Superior Técnico,
Universidade de Lisboa} \\ 
\and Anna Konovalenko \\\textit{ Molde University College, Norway}  } 
\begin{document}
\maketitle

\section{Introduction}


The emergence of large language models (LLMs) has produced a major advancement in natural language processing (NLP), catalyzing a fundamental transformation in information acquisition methodologies \citep{zhao2024surveylargelanguagemodels}. This breakthrough has led to substantial improvements in various areas of NLP, such as language generation, understanding and reasoning \citep{LeiMa25}. Furthermore, the extensive factual knowledge encoded within LLMs has shown significant progress in utilizing LLMs as knowledge bases \citep{alkhamissi2022reviewlanguagemodelsknowledge}, transforming the domain of information retrieval systems \citep{zhu2024largelanguagemodelsinformation}. However, with LLM's remarkable advancements, a major drawback is their tendency to hallucinate, producing responses that are misleading or entirely fabricated \citep{LeiMa25, guerreiro2023}. These hallucinations are particularly challenging to detect because LLMs generate sophisticated, human-like responses \citep{sadasivan2025}. The factually incorrect outputs from these models can significantly impact real-world decisions, potentially propagating misinformation and leading to harmful consequences in various contexts.  For example, when LLMs are used in sensitive or critical domains such as medicine, law, and finance, hallucinations can lead to severe errors, resulting in financial losses, diminished trust, and reputational damage.

Recent research has identified multiple strategies for reducing hallucination rates in LLMs and other generative architectures. One widely adopted method is Retrieval-Augmented Generation (RAG), which supplements the model’s parametric memory with external, verified knowledge repositories (e.g., curated databases, vector search indices, or real-time APIs). By constraining the generative process to draw from authoritative sources, RAG reduces reliance on potentially outdated or incomplete training data, thereby decreasing the probability of fabricating unsupported claims.

Post-generation verification constitutes another prominent mitigation paradigm. This approach employs secondary evaluator models, symbolic reasoning modules, or domain-specific rule-based checkers to assess the factual accuracy, logical consistency, and relevance of outputs prior to dissemination. Such evaluators operate as quality-control layers, filtering or revising responses before they reach end users.

Real-time hallucination detection has emerged as a distinct research trajectory, aiming to identify inaccuracies during the generation process itself. For example, the MIND framework \citep{su-etal-2024-unsupervised} leverages internal model state monitoring to detect anomalous activation patterns indicative of fabrication, circumventing the need for extensive labeled datasets. Similarly, Oxford’s semantic-entropy algorithm \cite{farquhar2024detecting} quantifies distributional variability across multiple sampled outputs for the same prompt, flagging high-variance responses as likely hallucinatory; this method has demonstrated detection accuracies of approximately 79 percents in controlled experiments.

A more recent line of work involves multiplex agent pipelines, in which multiple specialized AI agents are orchestrated sequentially, with each agent tasked with reviewing, refining, or fact-checking the prior agent’s output. This layered verification structure improves detection coverage and has shown potential in domains requiring high precision, such as biomedical information retrieval and legal text synthesis.

Additional mitigation strategies span prompt engineering techniques designed to constrain model behavior, integration of confidence estimation mechanisms to communicate epistemic uncertainty, deployment of human-in-the-loop workflows for high-risk or high-stakes contexts, and adversarial robustness measures to mitigate manipulation via crafted prompts.

Despite these advancements, a salient limitation in the current landscape is the absence of systematic hallucination classification. Contemporary mitigation systems primarily focus on binary detection—identifying whether a hallucination has occurred—without categorizing the type, cause, or severity of the inaccuracy. The lack of such classification constrains downstream handling: in high-risk applications, misclassifying a severe factual fabrication as a low-impact error may lead to insufficient intervention. Incorporating structured classification could enable risk-aware routing (e.g., escalating critical outputs for human review), targeted remediation tailored to the hallucination subtype (e.g., factual vs. logical inconsistency), and improved transparency through user-facing reliability labels. From a research perspective, classification would also facilitate the systematic study of hallucination patterns, enabling model developers to address root causes with greater precision.

The current level of automation and theorization in this direction remains low.  Most existing approaches to identifying and defining hallucinations are heavily based on manual methods, which negatively impact accuracy and reliability. Furthermore, LLMs can generate responses based on incomplete or indirect data, increasing the risk of errors and misleading information. The lack of a comprehensive formal approach to the detection of hallucinations complicates risk assessment and informed decision-making about model deployment. Over time, these risks can significantly limit the adoption of LLM, further underscoring the importance of this research.

Classifiers are essential in hallucination detection because they enable the formalization of the task as a supervised learning problem, allowing systematic evaluation and optimization. In research contexts, classifiers serve as decision functions that map model outputs to binary or multi-class labels (e.g., factual vs. hallucinated), facilitating reproducible benchmarking across datasets and model architectures. They can incorporate a range of features, from semantic similarity scores between generated text and reference sources to deeper representations learned by large language models themselves. By leveraging classifiers, researchers can quantify hallucination prevalence, compare mitigation strategies, and identify error patterns at scale, which is critical for advancing the scientific understanding of hallucination phenomena. Moreover, well-designed classifiers can generalize across domains and model types, supporting the development of standardized evaluation protocols in the broader hallucination detection literature.

In this work, we introduce a novel method for hallucination classification that involves generating diverse types of hallucinations and projecting the resulting outputs into a low-dimensional representation space. To the best of our knowledge, this is the first study to explore this approach.




Most current systems focus primarily on detection rather than detailed classification, but there are emerging approaches that do attempt to categorize different types of hallucinations.

The main contributions of this paper can be summarised as follows.
\begin{enumerate}
\item  the analysis of the current hallucination approaches and their limitations
\item proposing a novel framework for classifying hallucination
\item demonstate its success in classes generation
\end{enumerate}

\section{Background}
Hallucinations occur when models generate text that is unfaithful to source materials or factually incorrect. The underlying causes remain elusive due to the interplay between factors related to different stages of model development: data issues where models absorb biases and knowledge gaps; training problems that create patterns favoring fluency over accuracy; and inference challenges where decoding strategies and reasoning limitations produce plausible but incorrect outputs \citep{LeiMa25}. However, identifying hallucinations is important to improve model reliability and build systems that users can trust with critical information tasks.

Hallucination detection methods have emerged as a research area for identifying and quantifying these errors in model outputs. Hallucinations in language models can be classified into two fundamental categories: factuality hallucination, which occurs when generated content contradicts established real-world knowledge, and faithfulness hallucination, which emerges when outputs deviate from or misrepresent given source materials or instructions.

\textbf{Factuality:}
\begin{enumerate}
    \item Factual Contradiction (Inaccuracy) - when models generate content that contains objectively incorrect facts.
    \item Fabrication - when models invent non-existent information, entities, or events that contradict real-world knowledge.

\end{enumerate}

\textbf{Faithfulness:}
\begin{enumerate}
    \item A misinterpretation (Instruction Inconsistency for sure, Context Inconsistency as well) is a type of error where a language model misunderstands the context or the query itself, resulting in an incorrect response. Importantly, such a response is not necessarily based on factually incorrect information, but it does not correspond to the essence of the query or the content of the context.
    
    \item Context Inconsistency combines elements of Misinterpretation(if we leave context in misinterpretation) and Needle in a Haystack. A "needle in a haystack" is a situation where a language model is faced with a huge amount of data or information and is unable to extract the necessary details, resulting in partial inaccuracies or incomplete answers. This type of error often occurs when a model cannot effectively filter or analyze relevant information, and instead of providing an accurate answer to a query, it provides general or fragmentary information.
    (context either a separate and need is separate or context somewhere insert intro one of the classes)
    
    \item Structural hallucinations is more general than logical inconsistency (I prefer logical inconsistency) are a specific type of error that occurs when language models operate, where information is generated based on structural relationships rather than factual knowledge. These errors can reflect incorrect logical conclusions or distorted data, leading to distorted results.

\end{enumerate}


\section{Problem Description}

Detecting hallucinations in large language models remains a persistent and unresolved challenge. A core issue is the absence of reliable ground truth - for many prompts, especially open-ended or creative ones, there is no single correct answer, making it difficult to determine whether the model is hallucinating or being inventive. Another major difficulty lies in how convincing hallucinations can be; language models produce fluent and coherent text even when generating false information, making automatic detection tricky and human evaluation labor-intensive and inconsistent.

The goal of the solution method is to analyze the potential of distinguishing between correct answers and hallucinated ones, as well as to explore whether different types of hallucinations can be classified based on their semantic characteristics.

Existing approaches suffer from fundamental shortcomings: (1) they lack formal mathematical foundations and rely heavily on contextual information, (2) verification mechanisms are context-dependent and non-generalizable, (3) hallucination quality cannot be mathematically assessed, (4) current methods are computationally expensive and unreliable, often requiring one model to verify another, and (5) approaches are model-specific, preventing development of universal solutions.

We propose a formal approach grounded in linguistic theory, mapping ontology in language to separable surfaces. Our method exploits the modality gap to generate controlled hallucinations and establishes formal mathematical criteria for detection. We propose that hallucinations form distinct classes in representation space, enabling context-independent classification where truth statements and hallucinations constitute separate classes without requiring contextual knowledge.


\section{Solution method}

We propose a systematic approach to detect and analyze LLM hallucinations through geometric cluster analysis in the embedding space. Our methodology leverages the hypothesis that correct and hallucinated responses exhibit distinct distributional patterns when mapped to high-dimensional vector representations.

To accomplish our goal, we have designed a solution method which consists of the following stages:

\begin{enumerate}

\item \textbf{Data preparation}. Selection of question-answer datasets and generation of both correct and hallucinated responses using LLMs.

\item \textbf{Embedding conversion}. Transformation of all responses into high-dimensional vector embeddings for numerical representation.


\item \textbf{Dimensionality reduction and visualization}. Application of a dimensionality reduction technique to enable effective visualization of data distribution patterns.

\item \textbf{Clustering}. Construction of separate clusters for correct and hallucinated responses, with centroid computation across multiple runs for robust estimation.



\item \textbf{Quantitative analysis}. Measurement of distances between cluster centroids to quantify geometric separation between response types.


\end{enumerate}

\subsection{Data Preparation}

The dataset was sourced from open data \cite{naturalquestions_kaggle} but customized for this study to ensure reliable model evaluation. Several preprocessing steps were applied to clean and standardize the data.

The preprocessing pipeline involved common-use steps like text cleaning, remove unnecessary characters, HTML tags etc, followed by case normalization to convert text to lowercase and avoid duplication. The text was subsequently tokenized to facilitate accurate model interpretation, and duplicates—whether in queries submitted to the LLM or in its generated responses—were removed to mitigate potential sources of bias. Answer length filtering selected only questions with responses containing between min and max words. This focused the analysis on medium-length texts that are informative yet concise, avoiding extremes that could distort results.

After cleaning the dataset to remove irrelevant or corrupted entries, we employed the Llama3.1 \url{https://ollama.com/library/llama3.1}, Gemma2 \url{https://ollama.com/library/gemma2} and Phi3 \url{https://ollama.com/library/phi3} to generate both correct and hallucinated responses for each query, using corresponding prompt-engineering approaches.

\subsection{Embedding conversion }

All responses were transformed into high-dimensional vector embeddings to enable numerical analysis.
The all-MiniLM-L6-v2 model \url{https://huggingface.co/sentence-transformers/all-MiniLM-L6-v2} was then used to convert all responses into vector embeddings, maintaining consistency in the representation space. This approach ensured that both factual and hallucinated content were encoded using identical parameters to ensure that differences in the resulting embeddings are due solely to the content of the responses, not to variations in the encoding process, making comparisons reliable.

To evaluate the embeddings' ability to distinguish between response types, we calculated the average distances between the cluster centroids of model-generated answers and reference answers. Smaller distances indicated more accurate responses, whereas larger distances suggested hallucinated content. Multiple experimental rounds with different random seeds confirmed the stability of these patterns across various initial conditions.

Random seeds were systematically applied to ensure reproducible and robust results across different initial conditions and data splits, rather than being tied to any single fixed configuration.

Generated responses — with correct answers and hallucinations — from each model are processed independently and then mapped to a low-dimensional space.

\subsection{Dimensionality reduction and visualization}

UMAP (Uniform Manifold Approximation and Projection) \citep{mcinnes2018umap} was employed to reduce the dimensionality of the embeddings, making it easier to analyze and visualize their spatial properties. By projecting the high-dimensional data into a lower-dimensional space, UMAP simplifies the visualization and helps to better understand how embeddings are distributed. This approach enables clear comparison between model-generated answers and the correct references, as clusters of accurate responses and hallucinations become more distinguishable (this is a result, we expect it not that we now it at this stage). Additionally, by adjusting parameters such as $n_{neighbors}$, $min_{dist}$, and $random_{state}$, The analysis may investigate the influence of varying parameter settings on model behavior across diverse contexts by systematically monitoring cluster morphologies and inter-centroid distances.

\subsection{Quantitative analysis}

After reducing the dimensionality of the embeddings, each embedding was projected into a Cartesian coordinate system. This enabled the formation of spatial clusters corresponding to three response types: correct model-generated answers, ground truth answers from the dataset, and hallucinated outputs. The primary goal of the experiment was to analyze the spatial relationships between these groups and assess the model’s ability to align with ground truth or diverge toward hallucinations.

For each cluster, the centroid was computed. The centroid of a cluster is the mean of all data point coordinates within that cluster. Once centroids were computed, we measured the Euclidean distance between them. 

(the following paragraph is already a result, so not here)
The key finding was that the distance between the centroid of the correct model responses and the centroid of the ground truth was significantly smaller compared to the distance between the hallucinated responses' centroid and the ground truth. This indicates higher semantic similarity between correct and ground truth answers.

\section{Experimental setup}

All experiments were conducted on a MacBook Pro equipped with an Apple M2 chip and 16GB of RAM. The setup ensured consistent local testing conditions without reliance on external compute infrastructure. To ensure diversity and robustness in the experiments, three dataset sizes were used:

$n_1$ = 100 – small dataset for quick baseline results with minimal computational cost.

$n_2$ = 300 – medium dataset for more in-depth analysis while keeping processing time manageable.

$n_3$ = 1000 – large dataset for studying complex patterns and scalability effects.

The answers were generated using prompt engineering with constraints on the length of each response to ensure uniformity across outputs:

$L_{min}$ = 50 – minimum response length to maintain meaningful content.

$L_{max}$ = 70 – maximum response length to avoid overly verbose outputs.

For dimensionality reduction and embedding visualization, UMAP was employed with the following hyperparameters:

$k$ = 10 – number of nearest neighbors considered in the local manifold approximation.

$d_{min}$ = 0.2 – minimum distance between embedded points, controlling how tightly points are clustered.

$rs_u$ = 17 – random seed for reproducibility of UMAP results.

$\alpha$ = 0.8 – learning rate parameter influencing convergence speed.

$ms$ = 10 – minimum number of points in a local neighborhood for density estimation.

$\zeta$ = 1.2 – spread parameter affecting the overall scale of the embedding layout.



To test robustness and variability of the projections, each experiment was repeated using different $rs$ values ranging from 50 to 200 in steps of 50. This allowed us to measure the stability of UMAP-based centroid positioning across multiple initializations. Each experiment operated on a dataset of 500 question-answer pairs. The three answer groups included:

\begin{itemize}
    \item Ground truth answers from the Natural Questions Dataset
    \item Correct LLM-generated answers that match ground truth
    \item Hallucinated answers fabricated by LLM
\end{itemize}

For each experiment, centroids of the answer clusters were computed in the 2D UMAP space. Euclidean distances between these centroids were then calculated to assess semantic proximity. Results across four random seeds (50, 100, 150, 200) demonstrated consistent findings:

\begin{itemize}
    \item The distance between true answers and correct LLM answers was significantly smaller compared to distances involving hallucinated answers.
    \item This suggests that the LLM's correct outputs semantically cluster closer to ground truth, while hallucinations occupy a more distinct region in embedding space.
\end{itemize}

The complete implementation framework is publicly available on GitHub \citep{framework2024}, with visualization, clustering, centroid calculations, and analysis code accessible in the accompanying Jupyter notebook \citep{notebook2024}.

\section{Experimental Steps and Experimental Results}

To provide the analysis, we prepared and processed three types of datasets:

\begin{itemize}
    \item Ground-truth answers from the original dataset.
    \item Correct LLM-generated answers that match the ground-truth.
    \item Hallucinated LLM-generated answers containing factual inaccuracies.
\end{itemize}

To enable visual analysis and comparison, dimensionality reduction was applied to the embeddings. This step allowed us to observe the distribution and separation of different answer types in a reduced space.

Cluster construction and centroid distance measurement were performed using llm-hallucinations-detector-framework. Clusters representing each answer type were analyzed by calculating the distances between their centroids.

As a result, we experimentally confirmed that the distance between the centroids of expected answers and correct LLM answers is smaller than the distance between expected answers and hallucinated LLM answers. This supports the conclusion that correct LLM responses are semantically closer to ground-truth answers than hallucinated outputs. 

The following tables present the average distances between centroids for different categories of responses (correct answers and hallucinations) across various random seed initializations. The experiments were conducted with 500 samples and 3-dimensional UMAP projections, repeated over step sizes of 50, 100, 150, and 200 to assess consistency and robustness.

\begin{table}[h!]
\centering
\caption{Centroids average distances for different random seed initializations. Steps = 50. Test data shape = (500, 3)}
\begin{tabular}{|c|l|c|}
\hline
\textbf{Index} & \textbf{Key} & \textbf{Average Distance} \\
\hline
1 & NO TRUE expected answers, umap $\rightarrow$ correct\_answers, llama3\_1\_umap & 2.9272 \\
2 & NO TRUE expected answers, umap $\rightarrow$ hallucinations\_fabrication, llama3\_1\_umap & 5.9322 \\
3 & correct\_answers, llama3\_1\_umap $\rightarrow$ hallucinations\_fabrication, llama3\_1\_umap & 6.39826 \\
\hline
\end{tabular}
\end{table}

\vspace{0.5cm}

\begin{table}[h!]
\centering
\caption{Centroids average distances for different random seed initializations. Steps = 100. Test data shape = (500, 3)}
\begin{tabular}{|c|l|c|}
\hline
\textbf{Index} & \textbf{Key} & \textbf{Average Distance} \\
\hline
1 & NO TRUE expected answers, umap $\rightarrow$ correct\_answers, llama3\_1\_umap & 2.8951 \\
2 & NO TRUE expected answers, umap $\rightarrow$ hallucinations\_fabrication, llama3\_1\_umap & 5.826 \\
3 & correct\_answers, llama3\_1\_umap $\rightarrow$ hallucinations\_fabrication, llama3\_1\_umap & 6.4475 \\
\hline
\end{tabular}
\end{table}

\vspace{0.5cm}

\begin{table}[h!]
\centering
\caption{Centroids average distances for different random seed initializations. Steps = 150. Test data shape = (500, 3)}
\begin{tabular}{|c|l|c|}
\hline
\textbf{Index} & \textbf{Key} & \textbf{Average Distance} \\
\hline
1 & NO TRUE expected answers, umap $\rightarrow$ correct\_answers, llama3\_1\_umap & 2.39367 \\
2 & NO TRUE expected answers, umap $\rightarrow$ hallucinations\_fabrication, llama3\_1\_umap & 5.7424 \\
3 & correct\_answers, llama3\_1\_umap $\rightarrow$ hallucinations\_fabrication, llama3\_1\_umap & 6.3091 \\
\hline
\end{tabular}
\end{table}

\vspace{0.5cm}

\begin{table}[h!]
\centering
\caption{Centroids average distances for different random seed initializations. Steps = 200. Test data shape = (500, 3)}
\begin{tabular}{|c|l|c|}
\hline
\textbf{Index} & \textbf{Key} & \textbf{Average Distance} \\
\hline
1 & NO TRUE expected answers, umap $\rightarrow$ correct\_answers, llama3\_1\_umap & 2.9062 \\
2 & NO TRUE expected answers, umap $\rightarrow$ hallucinations\_fabrication, llama3\_1\_umap & 5.6738 \\
3 & correct\_answers, llama3\_1\_umap $\rightarrow$ hallucinations\_fabrication, llama3\_1\_umap & 6.2531 \\
\hline
\end{tabular}
\end{table}

The experimental results reveal several important patterns and insights related to the behavior of language models and their hallucination tendencies:

\begin{enumerate}
    \item \textbf{Centroid Distances:} The average Euclidean distances between centroids of hallucinated responses and other categories (true answers and correct LLM answers) are significantly higher than the distances between centroids of correct answers and true answers. This indicates a clear semantic separation in embedding space, where hallucinated responses form a distinct cluster.

    \item \textbf{Classification via Distance:} Based on these centroid distances, it is possible with high probability to classify whether a new response belongs to the category of hallucination or a correct answer. This provides a lightweight and interpretable mechanism for response validation.

    \item \textbf{Toward Fine-Grained Hallucination Typing:} Beyond binary classification, these distance-based features could enable classification of hallucination \emph{types}. This opens the door for further research where different forms of hallucinations (e.g., fabrication, distortion, omission) can be identified through clustering and distance patterns.

    \item \textbf{Reduced Computation for Detection:} This method significantly reduces the computational burden during inference, as it avoids the need for multiple LLM queries to validate a response. Embeddings and centroid-based calculations are lightweight and can be pre-computed or streamed efficiently.

    \item \textbf{Foundation for Hallucination Detection Model:} Most importantly, the experimental results lay a novel and promising foundation for training dedicated machine learning models that can detect hallucinations and even classify their types in real-time. These models can be built using supervised or semi-supervised learning on the embedding space, leveraging the structured and separable nature of hallucinated versus correct responses demonstrated in this study. This approach marks a new direction in hallucination detection—embedding-driven classification—shifting from reactive post-generation validation to proactive, efficient, and scalable inference-time hallucination detection.

\end{enumerate}

\begin{itemize}
    \item \href{https://www.kaggle.com/datasets/frankossai/natural-questions-dataset}{Natural Questions Dataset}    
\end{itemize}

\begin{figure}[h]
    \centering
    \includegraphics[width=0.7\textwidth]{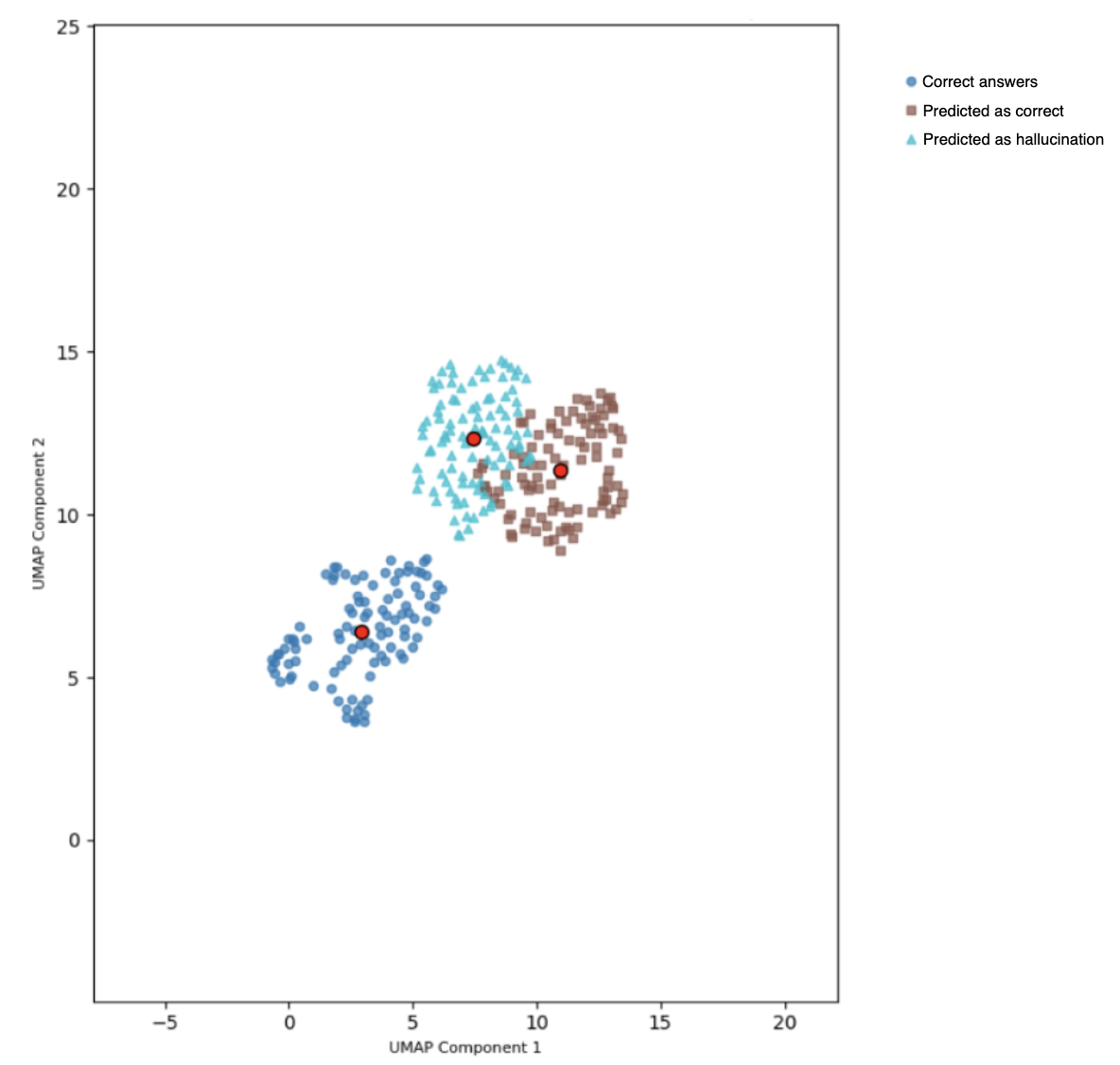}
    \caption{
    Visualization of sentence embeddings for 100 samples grouped into three clusters: (1) ground-truth correct answers, (2) LLM-generated answers predicted as correct, and (3) LLM-generated hallucinated answers. Each cluster represents a distinct category of responses. Red dots at the center of each cluster indicate the computed centroids. The visualization reflects the semantic grouping and separability of different answer types.
    }
    \label{fig:embedding_clusters_100}
\end{figure}

\begin{figure}[h]
    \centering
    \includegraphics[width=0.7\textwidth]{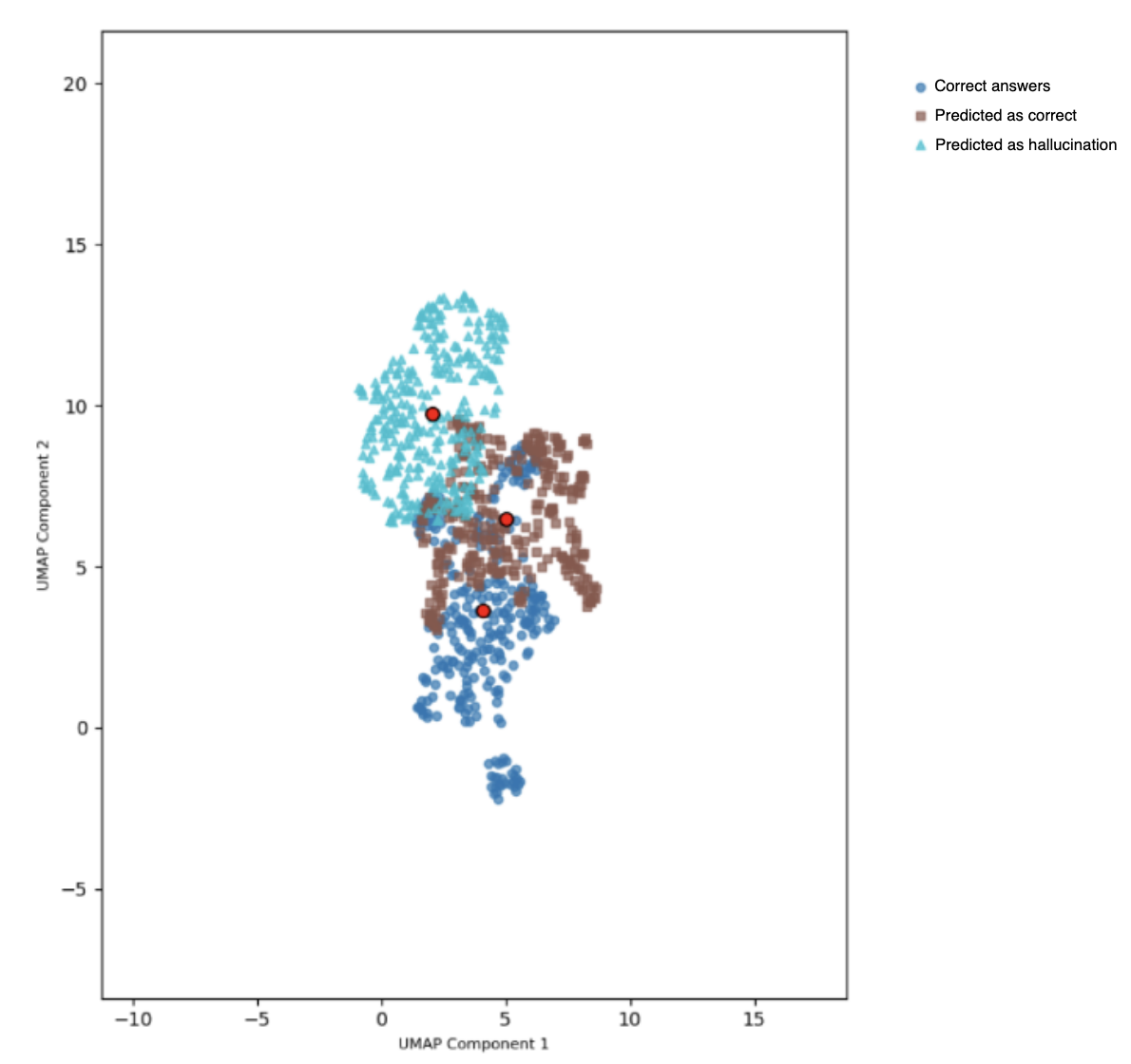}
    \caption{
    Visualization of sentence embeddings for 300 samples grouped into three clusters: (1) ground-truth correct answers, (2) LLM-generated answers predicted as correct, and (3) LLM-generated hallucinated answers. Each cluster represents a distinct category of responses. Red dots at the center of each cluster indicate the computed centroids. The visualization reflects the semantic grouping and separability of different answer types.
    }
    \label{fig:embedding_clusters_300}
\end{figure}

\begin{figure}[h]
    \centering
    \includegraphics[width=0.7\textwidth]{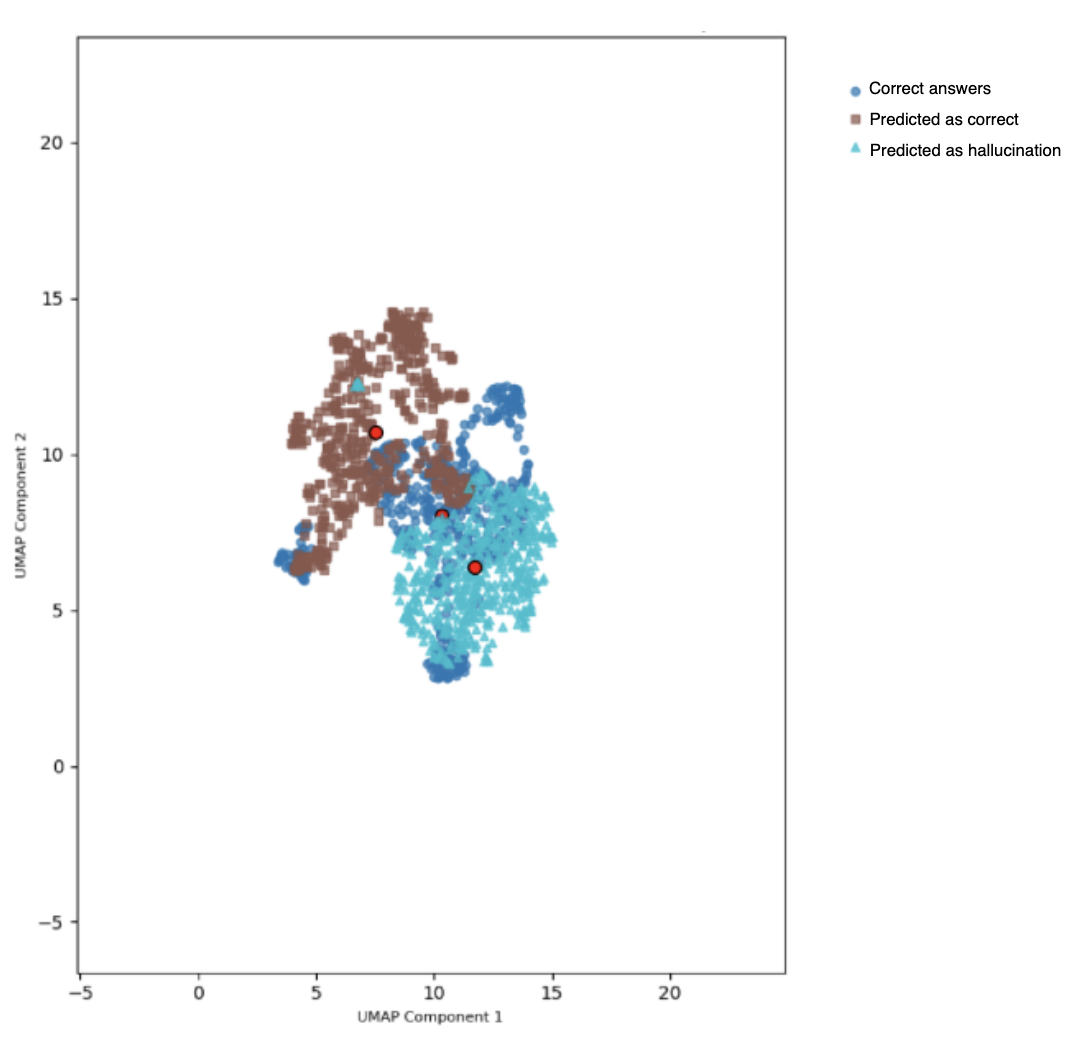}
    \caption{
    Visualization of sentence embeddings for 500 samples grouped into three clusters: (1) ground-truth correct answers, (2) LLM-generated answers predicted as correct, and (3) LLM-generated hallucinated answers. Each cluster represents a distinct category of responses. Red dots at the center of each cluster indicate the computed centroids. The visualization reflects the semantic grouping and separability of different answer types.
    }
    \label{fig:embedding_clusters_500}
\end{figure}

\clearpage

Each figure illustrates how sentence embeddings form distinct clusters corresponding to three categories of responses: ground-truth correct answers, LLM-generated answers predicted as correct, and LLM-generated hallucinations. These clusters were derived by embedding the textual data and projecting it into a reduced-dimensional space for visual interpretation. Responses from Llama3.1 model are shown.

The red dot at the center of each cluster indicates the centroid — the average position of all points within that group. The spatial separation between centroids reflects the semantic differences between answer types.

Across all three visualizations (100, 300, and 500 samples), the results consistently show that the centroid of the correct LLM answers is closer to the ground-truth centroid than the centroid of the hallucinated answers. This confirms that correct LLM outputs are semantically more aligned with reference answers, whereas hallucinations form a distinct and more distant cluster.

These findings experimentally support the effectiveness of embedding-based methods in identifying and separating hallucinated outputs from accurate LLM responses.

\section{Conclusion}
Our experiments demonstrated that hallucinations of different types are separable within a reduced-dimensional space. This separability was consistently reproducible across varying sample sizes and pseudorandom number generator parameters. The ability to construct a classifier that distinguishes between hallucination types, along with the capability to quantify distances between clusters of correct responses and hallucination clusters, provides a foundation for detecting hallucinations in a wide range of industrial applications. Furthermore, relating distance metrics in reduced-dimensional space to potential underlying ontological and semantic differences among hallucination types opens new avenues for deeper theoretical analysis.



\begin{thebibliography}{19}


\bibitem{zhao2024surveylargelanguagemodels}
Zhao, W.X., Zhou, K., Li, J., Tang, T., Wang, X., Hou, Y., Min, Y., Zhang, B., Zhang, J., Dong, Z., Du, Y., Yang, C., Chen, Y., Chen, Z., Jiang, J., Ren, R., Li, Y., Tang, X., Liu, Z., Liu, P., Nie, J.-Y., \& Wen, J.-R. (2024).
A Survey of Large Language Models.
arXiv preprint arXiv:2303.18223.
\url{https://arxiv.org/abs/2303.18223}

\bibitem{LeiMa25}
Huang, L., Yu, W., Ma, W., Zhong, W., Feng, Z., Wang, H., Chen, Q., Peng, W., Feng, X., Qin, B., \& Liu, T. (2025).
A Survey on Hallucination in Large Language Models: Principles, Taxonomy, Challenges, and Open Questions.
\textit{ACM Trans. Inf. Syst.}, 43(2), Article 42.
\url{https://doi.org/10.1145/3703155}

\bibitem{alkhamissi2022reviewlanguagemodelsknowledge}
AlKhamissi, B., Li, M., Celikyilmaz, A., Diab, M., \& Ghazvininejad, M. (2022).
A Review on Language Models as Knowledge Bases.
arXiv preprint arXiv:2204.06031.
\url{https://arxiv.org/abs/2204.06031}

\bibitem{zhu2024largelanguagemodelsinformation}
Zhu, Y., Yuan, H., Wang, S., Liu, J., Liu, W., Deng, C., Chen, H., Liu, Z., Dou, Z., \& Wen, J.-R. (2024).
Large Language Models for Information Retrieval: A Survey.
arXiv preprint arXiv:2308.07107.
\url{https://arxiv.org/abs/2308.07107}

\bibitem{guerreiro2023}
Guerreiro, N.M., Alves, D., Waldendorf, J., Haddow, B., Birch, A., Colombo, P., \& Martins, A.F.T. (2023).
Hallucinations in Large Multilingual Translation Models.
arXiv preprint arXiv:2303.16104.
\url{https://arxiv.org/abs/2303.16104}

\bibitem{sadasivan2025}
Sadasivan, V.S., Kumar, A., Balasubramanian, S., Wang, W., \& Feizi, S. (2025).
Can AI-Generated Text be Reliably Detected?
arXiv preprint arXiv:2303.11156.
\url{https://arxiv.org/abs/2303.11156}

\bibitem{su-etal-2024-unsupervised}
Su, W., Wang, C., Ai, Q., Hu, Y., Wu, Z., Zhou, Y., \& Liu, Y. (2024).
Unsupervised Real-Time Hallucination Detection based on the Internal States of Large Language Models.
In \textit{Findings of the Association for Computational Linguistics: ACL 2024} (pp. 14379--14391). Bangkok, Thailand.
\url{https://aclanthology.org/2024.findings-acl.854/}

\bibitem{farquhar2024detecting}
Farquhar, S., Kossen, J., Kuhn, L., et al. (2024).
Detecting hallucinations in large language models using semantic entropy.
\textit{Nature}, 630, 625--630.
\url{https://doi.org/10.1038/s41586-024-07421-0}

\bibitem{mcinnes2018umap}
McInnes, L., Healy, J., \& Melville, J. (2018).
UMAP: Uniform Manifold Approximation and Projection for Dimension Reduction.
arXiv preprint arXiv:1802.03426.

\bibitem{naturalquestions_kaggle}
Ossai, F. (2023).
Natural Questions Dataset.
Kaggle. \url{https://www.kaggle.com/datasets/frankossai/natural-questions-dataset}

\bibitem{framework2024}
Zavhorodnii, M. (2024).
LLM Hallucinations Detector Framework.
\url{https://github.com/imaxfp/llm-hallucinations-detector-framework}

\bibitem{notebook2024}
Zavhorodnii, M. (2024).
Structural Hallucinations Analysis Notebook.
\url{https://github.com/imaxfp/llm-hallucinations-detector-framework/blob/main/main1_structural_hallucinations.ipynb}

\end{thebibliography}

\end{document}